\documentclass[10pt,twocolumn,letterpaper]{article}

\usepackage{iccv}
\usepackage{times}
\usepackage{epsfig}
\usepackage{graphicx}
\usepackage{amsmath}
\usepackage{amssymb}
\usepackage[utf8]{inputenc}
\usepackage{multirow}
\usepackage{multicol}
\newcommand{\walon}[1]{{\color{black}#1}\normalfont}

\usepackage[breaklinks=true,bookmarks=false]{hyperref}

\iccvfinalcopy %

\def\iccvPaperID{1238} %

\ificcvfinal\fi

\begin{document}

\title{RPG: Learning Recursive Point Cloud Generation}

\author{Wei-Jan Ko\\

{\tt\small ts771164@gmail.com}
\and
Hui-Yu Huang\\

{\tt\small purelyvivid@gmail.com}

\and
Yu-Liang Kuo\\

{\tt\small hank.cs08g@nctu.edu.tw}

\and
Chen-Yi Chiu\\

{\tt\small charles.en07@nycu.edu.tw}

\and
Li-Heng Wang\\

{\tt\small leo.eecs06@nctu.edu.tw}

\and
Wei-Chen Chiu\\

{\tt\small walon@cs.nctu.edu.tw}

\and
National Chiao Tung University

}

\maketitle
\ificcvfinal\thispagestyle{empty}\fi

\begin{abstract}
\walon{In this paper we propose a novel point cloud generator that is able to reconstruct and generate 3D point clouds composed of semantic parts. Given a latent representation of the target 3D model, the generation starts from a single point and gets expanded recursively to produce the high-resolution point cloud via a sequence of point expansion stages. During the recursive procedure of generation, we not only obtain the coarse-to-fine point clouds for the target 3D model from every expansion stage, but also unsupervisedly discover the semantic segmentation of the target model according to the hierarchical/parent-child relation between the points across expansion stages. Moreover, the expansion modules and other elements used in our recursive generator are mostly sharing weights thus making the overall framework light and efficient. Extensive experiments are conducted to demonstrate that our proposed point cloud generator has comparable or even superior performance on both generation and reconstruction tasks in comparison to various baselines, as well as provides the consistent co-segmentation among 3D instances of the same object class.}
\end{abstract}

\vspace{-1em}
\section{Introduction}
\walon{
As the increasing usage of 3D sensors nowadays, understanding the 3D data and the rich geometric information of it becomes a crucial problem in many applications such as robotics and autonomous vehicles. Among various formats of 3D data (e.g. meshes, voxels, and point clouds), the point cloud is one of the most intuitive and popular representations, not only because of being the default output of several 3D sensors (e.g. LiDARs) but also its ability of well capturing the geometric details. In this paper, we focus on better understanding the underlying geometry of 3D data via learning the generative procedure of point clouds.
}

\walon{Without loss of generality, understanding the geometry of 3D point clouds can be categorized into three levels of granularity from fine to coarse, as sequentially described in the following: \textbf{(1)} \emph{Modelling structure details}, where many works \cite{Yang_2018_CVPR,Groueix_2018_CVPR,ShapeGF,Yang_2019_ICCV,hui2020pdgn,pmlr-v80-achlioptas18a} proposed for generating or reconstructing point clouds belongs to such level, in which various modelling techniques are adopted to describe the geometric details. For instance, \cite{Yang_2018_CVPR,Groueix_2018_CVPR} utilize the transformation of multiple 2D grids/patches to reconstruct the 3D object surface, while \cite{ShapeGF} builds the probabilistic model based on point densities for generating sophisticated surfaces through sampling; 
\textbf{(2)} \emph{Extracting semantic parts}, where a point cloud is represented as the (hierarchical) combination of semantic parts. For instance, both \cite{Hertz_2020_CVPR} and \cite{Shu_2019_ICCV} propose tree-structured generative models of point clouds, where the former explicitly constructs hierarchical part-based representations while the latter produces part segmentation of the generated point cloud via tracing the tree built upon graph convolutions;
and \textbf{(3)} \emph{Discovering shape correspondences}, where the co-analysis of multiple point clouds of the same class is applied to discover the correspondences across shapes. For example, \cite{Chen_2019_ICCV} learns the co-segmentation while \cite{Chen_2020_CVPR} finds the semantic-aware structural points shared across object instances.
}

\begin{figure}
\begin{center}
\includegraphics[width=\columnwidth]{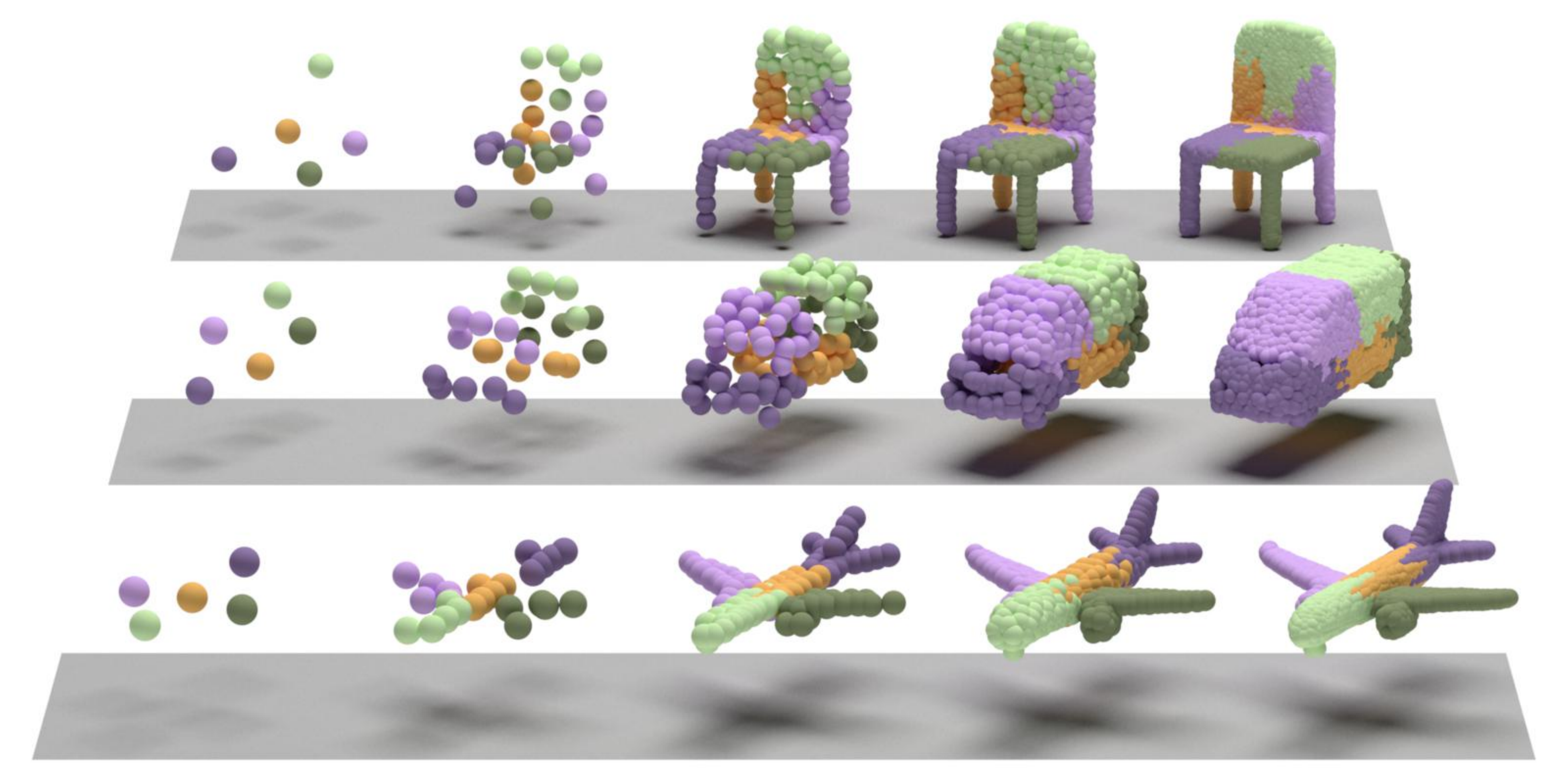}
\end{center}
   \caption{\walon{With the novel design of having recursive point expansion stages, our proposed model is able to not only produce coarse-to-fine generation of point clouds, but also unsupervisedly discover their semantic parts, which provides more fine-grained understanding on the geometry of 3D shapes.}}
\vspace{-1em}
\label{fig:teaser}
\end{figure}

\walon{While most of the research works of understanding the geometry of point clouds only focus on one of the aforementioned levels (with few of them taking care of two levels), in this paper we aim to model the generative procedure of point cloud data with including all the granularities in a unified framework. We propose a \textbf{recursive point cloud generator (RPG)} which takes a latent vector (standing for the representation of the whole target 3D shape) as input, and generates the point clouds in a coarse-to-fine manner via a sequence of point expansion stages, as shown in Figure~\ref{fig:teaser}. Basically, the generation starts from initializing the simplest/coarsest point cloud composed of only a single 3D point at origin, the expansion module at each expansion stage recursively duplicates every point obtained from the previous stage certain times to form a finer point cloud (i.e. having more points to detailedly represent the target 3D shape). The point clouds produced from all the stages naturally form a tree-structure, where we can partition the points at an arbitrary stage according to their ancestor assignments on a previous stage thus resulting the hierarchical part segmentation in an unsupervised way. Moreover, as the RPG is learnt on multiple 3D instances at the same time, the shape correspondences across point clouds can be automatically discovered. With having the expansion modules sharing weights across expansion stages by the recursive nature of our RPG, our proposed framework results to have small  model size thus leading to more efficient model training. We experimentally show that our RPG framework is able to achieve comparable generation and reconstruction performance with respect to the state-of-the-art baselines, and produce compact and semantically-meaningful part segmentation/co-segmentation on point clouds.}

\section{Related Work}
\walon{In the following we describe the related works for understanding geometry of 3D point cloud data from three granularities, as aforementioned in the introduction.}

\noindent\textbf{Reconstruction and Generation.}~
\walon{Though the popularity of using point cloud to represent 3D data, the tasks of reconstructing or generating the point cloud data are still considered to be challenging due to its irregularity (e.g. permutation-invariance). Various research works have been proposed these years, for instance:~\cite{Yang_2018_CVPR} and~\cite{Groueix_2018_CVPR} represent a 3D shape as the deformation of 2D grid points or the 2D parametric surface elements, which however cannot well reconstruct the local details of point cloud data; \cite{Yang_2019_ICCV} models the point clouds as a two-level hierarchy of distributions via adopting the frameworks based on continuous normalizing flow; \cite{hui2020pdgn} develops a progressive deconvolution network to generate the coarse-to-fine point clouds, which is trained via adversarial learning; and \cite{ShapeGF} proposes an energy-based probabilistic framework to model the distribution of 3D points via learning the gradient field of the point densities. While these prior works improve the fidelity of generated or reconstructed point clouds over the recent years, the information or the segmentation of semantic parts are not taken into consideration for their model designs.  
}

\noindent\textbf{Hierarchical and Part-based Representation.}~
\walon{Treating the 3D shape as a hierarchical decomposition of parts has facilitated humans' understanding on the shape structures, and thus have attracted plenty research attentions. For instance, \cite{mo2019structurenet} and \cite{mo2020structedit} represent 3D objects as hierarchical $n$-ary trees and train the graph neural network~\cite{kipf2016semi} and the conditional variational autoencoder respectively to learn the hierarchical structures, where they require the dataset with groundtruth part annotations (e.g. PartNet~\cite{mo2019partnet}) for the model learning. In contrast, Hertz \etal~\cite{Hertz_2020_CVPR} learns the hierarchical part-based representation of point clouds in an unsupervised manner, where they adopt the tree-structured Gaussian mixture distribution to probabilistically approximate the point cloud surface. However, their model struggles to produce small sharp details in the generated point cloud, as indicated by themselves in \cite{Hertz_2020_CVPR}. Similarly, TreeGAN proposed by~\cite{Shu_2019_ICCV} learns a point cloud generator built upon tree-structured graph convolution network, but only outputs the point cloud with part segmentation at the final layer and lacks of the intermediate point clouds during the generation, i.e. having no hierarchical representation.}

\noindent\textbf{Cross-Object Shape Correspondence.}~
\walon{Discovering the semantic correspondences across 3D object instances helps to investigate and study the variances among 3D shapes thus leading to more high-level understanding. \cite{Genova_2019_ICCV,Paschalidou2019CVPR,Tulsiani_2017_CVPR} learn the decomposition of shapes with adopting the primitive-based representation, where the parts are however limited to the primitives such as cuboids, superquadrics, or convex hulls. \cite{Chen_2020_CVPR} instead proposes an unsupervised method to discover the 3D structural points which are shared among the shape instances with similar structures and show the semantic consistency. And \cite{Chen_2019_ICCV} leverages a branched autoencoder to achieve co-segmentation between similar shapes, where each branch represents a recurring part in the implicit field. Although producing semantic correspondences across shapes, \cite{Chen_2020_CVPR,Chen_2019_ICCV} do not support the point cloud generation.}

\walon{In contrast to all the aforementioned works, our proposed recursive point cloud generator (RPG) is able to perform reconstruction and generation of point cloud data with high fidelity, produce hierarchical part-based segmentation without requiring any supervision on part annotations, and discover the semantic correspondence across instances of the same category, with all achieved by a lightweight model.}

\begin{figure*}[ht!]
    \centering
    \includegraphics[width=.9\textwidth]{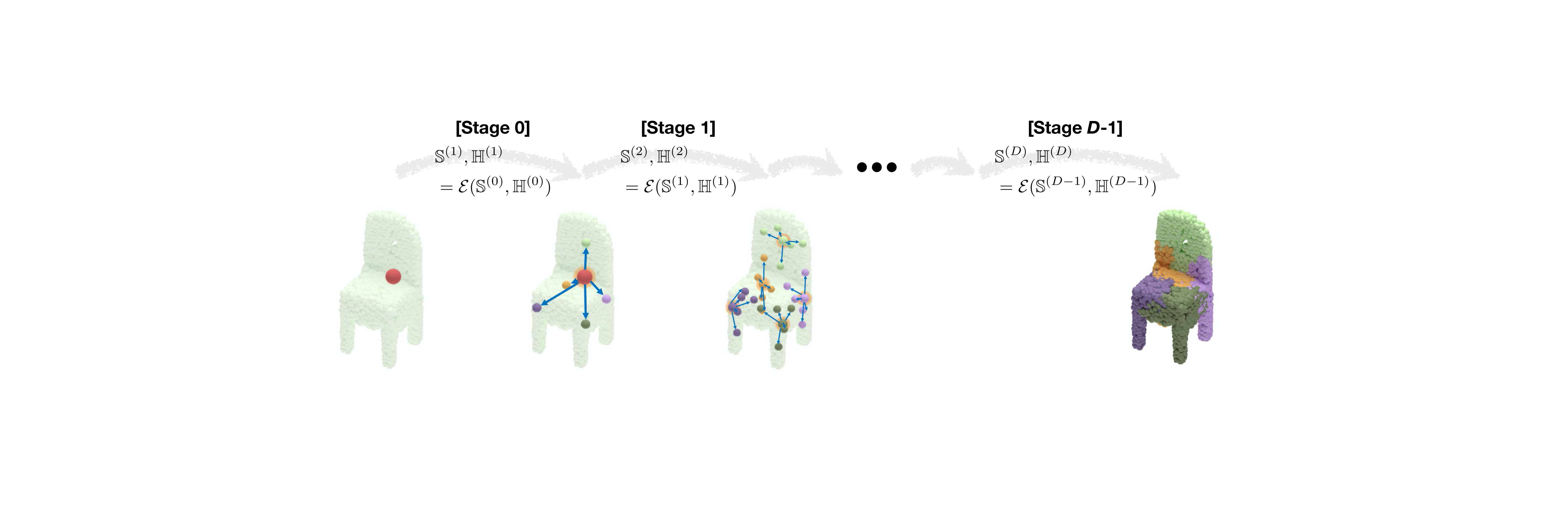}
    \caption{\walon{Illustration for the basic concept behind our proposed Recursive Point Cloud Generator (RPG). Assuming now we are aiming to reconstruct the point cloud of a 3D chair model, with being given its latent/structural representation $z$. Our generation starts from having $\mathbb{S}^{(0)} = \{\mathbf{0}\}$ and $\mathbb{H}^{(0)} = \{z\}$. The expansion module $\mathcal{E}$ in the Stage 0 firstly expands from the origin point $\mathbf{0}$ into $k(0)=5$ points $\mathbb{S}^{(1)} = \{s^{(1)}_{i}\}_{i=1}^{k(0)}$, which are roughly distributed to the locations of the chair back and chair legs, as well as evolves $z$ into $k(0)$ structural representations $\mathbb{H}^{(1)} = \{h^{(1)}_{i}\}_{i=1}^{k(0)}$ indicating that now the chair model is assembled by $k(0)$ parts. The expansion module $\mathcal{E}$ in the Stage 1 can further expand every point in $\mathbb{S}^{(1)}$ and representation in $\mathbb{H}^{(1)}$ by $k(1)$ times in order to derive $\mathbb{S}^{(2)}$ and $\mathbb{H}^{(2)}$, where now the point cloud for the chair contains more points thus becoming more fine-grained. With having the expansion module recursively applied $D$ times, we obtain $\mathbb{S}^{(D)}$ as the resultant point cloud for the 3D chair model, in which there are $\prod_{d=0}^{D-1} k(d)$ points in the point cloud. The overall procedure of our RPG shows the coarse-to-fine generation of the 3D point cloud, and we can easily obtain the part segmentation of the point cloud according to the  corresponding ancestors (e.g. in $\mathbb{S}^{(1)}$) for each point in $\mathbb{S}^{(D)}$, as shown in the right-most figure.}
    } 
    \vspace{-1em}
    \label{fig:illustration}
\end{figure*}

\section{Recursive Point Cloud Generator (RPG)}
\walon{The goal of our proposed method is to unsupervisedly learn a point cloud generator, named as \textbf{Recursive Point Cloud Generator} (\textbf{RPG}), which is able to generate 3D point clouds in a recursive coarse-to-fine manner. The learning of RPG is based on the autoencoder framework: Given a point cloud data $\mathcal{P}$, an encoder $E$ (which is built upon the PointNet~\cite{Qi_2017_CVPR} followed by a fully-connected layer in our implementation) first maps it into a latent representation $z \in \mathbb{R}^{U}$, our proposed RPG then takes $z$ as input and attempts to reconstruct $\hat{\mathcal{P}}$ which ideally should be identical to $\mathcal{P}$. The reconstruction error between $\hat{\mathcal{P}}$ and $\mathcal{P}$ is thus the objective to drive the training of encoder $E$ and our RPG.}

\walon{Now we detail the generative process of our proposed RPG for generating the point cloud data from the latent representation $z$. Basically, we decompose the generative process into $D$ sequential stages of point expansion, as formulated in Eq.~\ref{eq:whole_expansion}. At each stage $d$, the 3D points $\mathbb{S}^{(d)}$ together with their corresponding structural representations $\mathbb{H}^{(d)}$ are expanded by $k(d)$ times into $\mathbb{S}^{(d+1)}$ and $\mathbb{H}^{(d+1)}$ respectively via the expansion module $\mathcal{E}$, where $d=0, 1, \cdots, D-1$. In other words, the number of 3D points in $\mathbb{S}^{(d+1)}$, denoted as $\left| \mathbb{S}^{(d+1)} \right|$, is $k(d)$ times more than the one in $\mathbb{S}^{(d)}$. The 3D points obtained after the last stage of expansion, i.e. $\mathbb{S}^{(D)}$, become the output $\hat{\mathcal{P}}$ of our RPG, in which $\left| \mathbb{S}^{(D)} \right|$ can be also calculated easily by $\prod_{d=0}^{D-1} k(d)$.}

\begin{equation}
\begin{aligned}
\text{Stage~0:~~}&\quad \mathbb{S}^{(1)}, \mathbb{H}^{(1)} = \mathcal{E} (\mathbb{S}^{(0)}, \mathbb{H}^{(0)})\\
\text{Stage~1:~~}&\quad \mathbb{S}^{(2)}, \mathbb{H}^{(2)} = \mathcal{E}(\mathbb{S}^{(1)}, \mathbb{H}^{(1)})\\
&\qquad\small{\vdots}\\
\text{Stage~$D-1$:}&\quad \mathbb{S}^{(D)}, \mathbb{H}^{(D)} = \mathcal{E}(\mathbb{S}^{(D-1)}, \mathbb{H}^{(D-1)}).
\end{aligned}
\label{eq:whole_expansion}
\end{equation}

\walon{The basic concept behind these sequential stages can be described in an intuitive way, as illustrated in Figure~\ref{fig:illustration}: Assuming that we are now asked to build the 3D point cloud of a specific chair, but at the beginning we can only use $5$ points to represent the structure of that chair. It then become a reasonable choice to distribute these $5$ points to the chair back and $4$ chair legs respectively. Such scenario is analogous to our stage 0 of point expansion with taking $\mathbb{H}^{(0)} = \{z\}$ and $\mathbb{S}^{(0)} = \{ \mathbf{0} \}$, where the $k(0)=5$ points (denoted as $\mathbb{S}^{(1)} = \{s^{(1)}_{i}\}^{k(0)}_{i=1}$) are expanded from the origin $\mathbf{0}$ according to the latent structural representation $z$ of the whole chair, and $z$ is evolved into $\mathbb{H}^{(1)} = \{h^{(1)}_{i}\}^{k(0)}_{i=1}$ representations indicating that now the chair is assembled by $k(0)=5$ parts. When stepping further with being allowed to use more points to represent each part of this chair, these points would ideally be distributed to better cover the volume of their corresponding parts. Similarly, our stage 1 is analogous to this scenario, in which for every point $s^{(1)}_i \in \mathbb{S}^{(1)}$ we can expand it into $k(1)$ more points as well as evolve its structural representation $h^{(1)}_i \in \mathbb{H}^{(1)}$ into $k(1)$ corresponding structural representations, indicating that now the part related to $s^{(1)}_i$ is represented by $k(1)$ more sub-parts. All the points and structural representations expanded from $\mathbb{S}^{(1)}$ and $\mathbb{H}^{(1)}$ are denoted as $\mathbb{S}^{(2)}$ and $\mathbb{H}^{(2)}$ respectively, in which $\mathbb{S}^{(2)}$ shows more finer-grained 3D point cloud of the chair in comparison to $\mathbb{S}^{(1)}$. Following the similar principle as previous stages, more stages of point expansion can be recursively applied to build the 3D point cloud in a coarse-to-fine manner.}

\subsection{Expansion Module $\mathcal{E}$}\label{sec:expansion}
\begin{figure*}[ht!]
    \centering
    \includegraphics[width=\textwidth]{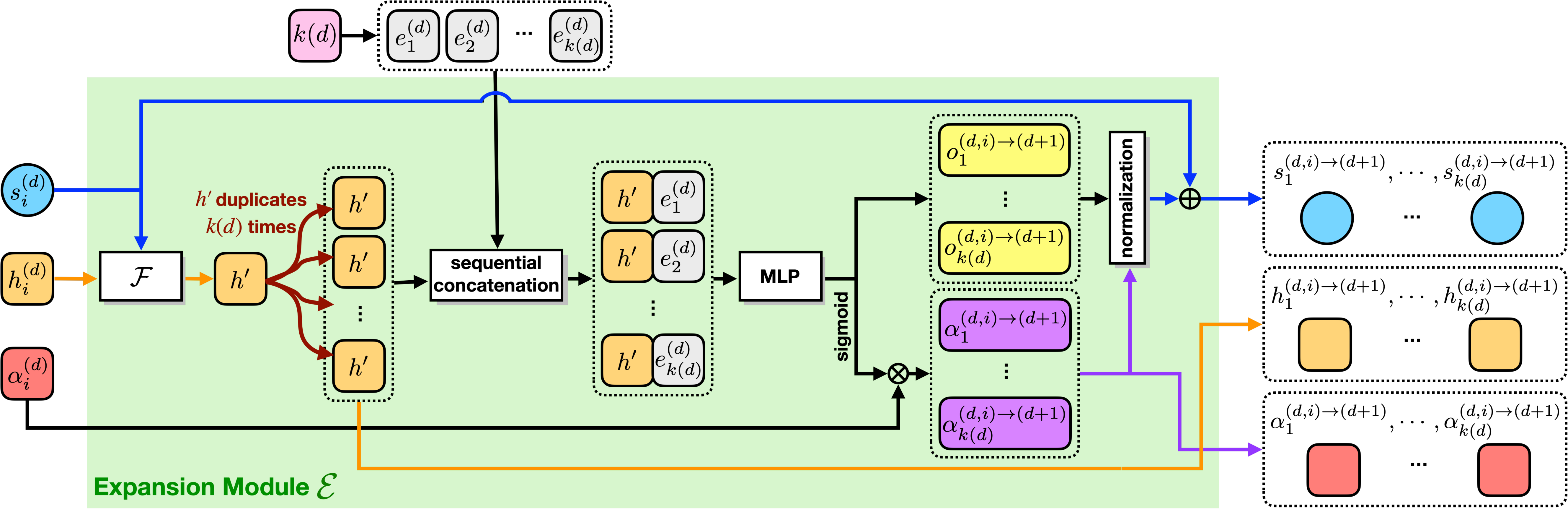}
    \caption{Architecture of our expansion module $\mathcal{E}$. Please refer to Section~\ref{sec:expansion} for more detailed description.}
    \label{fig:expansion}
\end{figure*}

\walon{The architecture of the expansion module $\mathcal{E}$ used in RPG is illustrated in Figure~\ref{fig:expansion}. Given a 3D point $s_i^{(d)}\in \mathbb{S}^{(d)}$ with its corresponding structural representation $h_i^{(d)}$ and scaling factor $\alpha_i^{(d)}$, the expansion module $\mathcal{E}$ expands $s_i^{(d)}$, $h_i^{(d)}$, and $\alpha_i^{(d)}$ respectively by $k(d)$ times:}
\begin{equation}
\begin{aligned}
&\left \{ s^{(d,i)\rightarrow(d+1)}_{m}, h^{(d,i)\rightarrow(d+1)}_{m}, 
\alpha^{(d,i)\rightarrow(d+1)}_{m} \right \}^{k(d)}_{m=1}\\
&\qquad\qquad\qquad\qquad\qquad\qquad= \mathcal{E}(s^{(d)}_i, h^{(d)}_i, \alpha^{(d)}_i)
\end{aligned}
\label{eq:expansion_module} 
\end{equation}
\walon{where $h^{(d,i)\rightarrow(d+1)}_{1} = h^{(d,i)\rightarrow(d+1)}_{2} =\cdots=h^{(d,i)\rightarrow(d+1)}_{k(d)}$}.
\walon{Please note that in Eq.~\ref{eq:whole_expansion} we skip the scaling factors $\alpha$ for simplicity. With denoting all the 3D points and structural representations expanded from $s_i^{(d)}$ and $h_i^{(d)}$ respectively as $\mathbb{S}^{(d,i)\rightarrow(d+1)}$ and $\mathbb{H}^{(d,i)\rightarrow(d+1)}$, the overall points and their corresponding structural representations obtained at the stage $d$ are then accumulated respectively by:}
\begin{equation}
    \begin{aligned}
    \mathbb{S}^{(d+1)} &= \bigcup\limits_{i=1}^{\left| \mathbb{S}^{(d)} \right|} \mathbb{S}^{(d,i)\rightarrow(d+1)}\\
    \mathbb{H}^{(d+1)} &= \bigcup\limits_{i=1}^{\left| \mathbb{H}^{(d)} \right|} \mathbb{H}^{(d,i)\rightarrow(d+1)}
    \end{aligned}
\end{equation}

\walon{We now provide more detailed explanation for how such expansion module works as well as the physical meanings behind each variable. We say, the parent point of $s^{(d)}_i$ (denoted as $P(s^{(d)}_i) \in \mathbb{S}^{(d-1)}$, which $s^{(d)}_i$ is expanded from) is in charge of representing a certain portion $\mathcal{V}$ of the target 3D model (e.g. a 3D chair). According to Eq.~\ref{eq:expansion_module}, all the sibling points of $s^{(d)}_i$, which are also expanded from $s^{(d)}_i$, do share the same structural representation $h_i^{(d)}$ as $s^{(d)}_i$. Therefore, $h_i^{(d)}$ actually carries the structural information of $\mathcal{V}$. Now, when we would like to further expand $s^{(d)}_i$ into $k(d)$ points, a function $\mathcal{F}$ is adopted to extract from $h_i^{(d)}$ the structural information ${h}'$ for the sub-portion $\mathcal{V}'$ of $\mathcal{V}$ that $s^{(d)}_i$ is taking care of.}
\begin{equation}
    {h}' = \mathcal{F}(s^{(d)}_i, h^{(d)}_i)
    = \text{tanh}(\mathcal{M}_{s}\ast s^{(d)}_i + \mathcal{M}_{h}\ast h^{(d)}_i)
\end{equation}
\walon{where $s^{(d)}_i \in \mathbb{R}^{3\times 1}$, $\mathcal{M}_{s} \in \mathbb{R}^{U\times 3}$, $h^{(d)}_i \in \mathbb{R}^{U\times 1}$, and $\mathcal{M}_{h} \in \mathbb{R}^{U\times U}$ ($U$ is set to $512$ in all our experiments). Then, as all the child points expanded from $s^{(d)}_i$ would share the same ${h}'$, we duplicate it $k(d)$ times, and denote them as $\mathbb{H}^{(d,i)\rightarrow(d+1)} = \{h^{(d,i)\rightarrow(d+1)}_{m}\}_{m=1}^{k(d)}$. Regarding the scaling factor $\alpha^{(d)}_i$ of the point $s^{(d)}_i$, it is to constraint the range of movement for $s^{(d)}_i$ when $s^{(d)}_i$ is expanded from its parent $P(s^{(d)}_i)$ as $\mathcal{V}$ is just a portion of the whole target 3D model, i.e. $s^{(d)}_i$ should not exceed the spatial range of $\mathcal{V}$.  
}

\walon{Regarding the way of distributing $k(d)$ points in the sub-portion $\mathcal{V}'$ that $s^{(d)}_i$ takes care of, we particularly introduce the embeddings $\{e^{(d)}_m\}_{m=1}^{k(d)}$ which serve as the high-level representations for an initial spatial-arrangement, the concatenation of each $e^{(d)}_m$ with ${h}'$ is then taken as the input for a multiple-layer perceptron (MLP) to produce the offset $o^{(d,i)\rightarrow(d+1)}_m \in \mathbb{R}^{3\times 1}$ and the scaling factor $\alpha^{(d,i)\rightarrow(d+1)}_m$ of the $m$-th point expanded from $s^{(d)}_i$. Please note that now the scaling factor $\alpha^{(d,i)\rightarrow(d+1)}_m$ has been multiplied with $\alpha^{(d)}_i$ after MLP. The resultant 3D point coordinate $s^{(d,i)\rightarrow(d+1)}_m$ of $m$-th point is computed by:}
\begin{equation}
    s^{(d,i)\rightarrow(d+1)}_m = s^{(d)}_i +  \frac{o^{(d,i)\rightarrow(d+1)}_{m}}{o_{\text{max}}} \ast \alpha^{(d,i)\rightarrow(d+1)}_m
\end{equation}
\walon{where $o_{\text{max}} = \max(\{ \left\|o^{(d,i)\rightarrow(d+1)}_{m} \right\| \}_{m=1}^{k(d)})$.}

\begin{figure}[h!]
    \centering
    \includegraphics[width=\columnwidth]{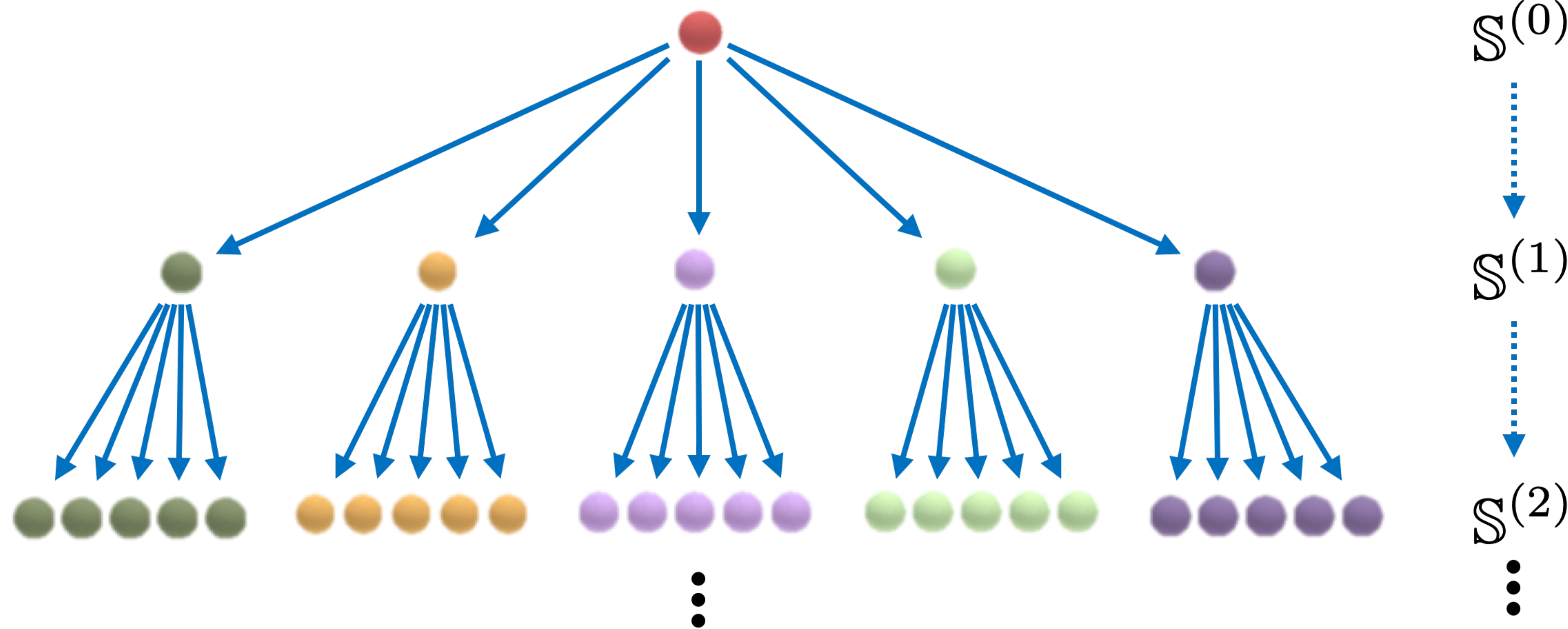}
    \caption{Illustration of the tree-structure built upon the point clouds obtained from the expansion stages of our RPG.}
    \vspace{-1em}
    \label{fig:tree}
\end{figure}

\walon{\textbf{Insights.~} As the points in $\mathbb{S}^{(d)}$ are expanded the ones in $\mathbb{S}^{(d-1)}$, with leveraging the hierarchical/parent-child relation between them, we can straightforwardly construct a tree-structure upon all the point clouds (i.e. $\{\mathbb{S}^{(d)}\}_{d=0}^{D-1}$) obtained during the process of our recursive point generator, as illustrated in Figure~\ref{fig:tree}. Note that, such tree-structure will form a perfect $k$-ary tree if all $k(d)$ are equivalent $\forall d$. Moreover, when we consider the path of tracing each point in $\mathbb{S}^{(D)}$ (i.e. leave node) back to the origin point in $\mathbb{S}^{(0)}$ (i.e. root node), the function $\mathcal{F}$ with respect to all the points and their structural representations in such path actually form a simplest recurrent neural network (RNN), in which the points $s$ and structural representations $h$ in the path are the inputs and hidden states respectively, while $\mathcal{M}_s$ and $\mathcal{M}_t$ are the parameters of such RNN.}

\walon{Lastly, when we consider the cluster assignments for all the points in $\mathbb{S}^{(d)}$ according to their corresponding ancestor points in $\mathbb{S}^{({d}')}$, where ${d}'<d$, we can obtain the segmentation result of $\mathbb{S}^{(d)}$ with $|\mathbb{S}^{({d}')}|$ clusters. Examples of using $\mathbb{S}^{(1)}$ as the reference of the ancestor points to perform segmentation on various $\mathbb{S}^{(d)}$ with $d>1$ are shown in Figure~\ref{fig:teaser}. 
}

\subsection{Training Objective}
\walon{The main objective for our RPG training simply adopts the Chamfer Distance (CD)~\cite{Barrow1977ijcai} for evaluating the reconstruction error between $\mathcal{P}$ and our RPG output $\hat{\mathcal{P}} = \mathbb{S}^{(D)}$.}
\begin{equation}
    \begin{aligned}
    \mathcal{L}_{\text{CD}} = \text{CD}(\mathcal{P},\hat{\mathcal{P}})&=\frac{1}{|\mathcal{P}|}\sum_{{s}\in {\mathcal{P}}}\min_{{s}'\in \hat{\mathcal{P}}}||s-{s}'||_2^2 \\ &+
    \frac{1}{|\hat{\mathcal{P}}|}\sum_{{s}'\in {\hat{\mathcal{P}}}}\min_{s\in \mathcal{P}}||s-{s}'||_2^2
    \end{aligned}
\end{equation}
\walon{
Moreover, we introduce a regularization term to generally encourage smaller scaling factors, in order to reduce the potential overlapping between different portions (i.e. $\mathbb{S}^{(d,i)\rightarrow(d+1)}$ and $\mathbb{S}^{(d,j)\rightarrow(d+1)}$, $i\neq j$) on each stage.}
\begin{equation}
    \mathcal{L}_{\text{reg}} = \frac{1}{D} \sum_{d=0}^{D-1}\left (\frac{1}{|\mathbb{S}^{(d)}|}\sum_{i=1}^{|\mathbb{S}^{(d)}|} \mathbb{A}^{(d)}\right )
\end{equation}
\walon{where $\mathbb{A}^{(d)}$ denotes all the scaling factors obtained at stage $d$. The overall objective then becomes $\mathcal{L}_{\text{CD}} + \lambda \mathcal{L}_{\text{reg}}$, where $\lambda$ is set to $5\mathrm{e}{-5}$ in our experiments.}

\walon{Please especially note that, the parameters of $\mathcal{F}$ (i.e. $\mathcal{M}_s$ and $\mathcal{M}_t$) and the ones of MLP are shared globally across all the stages of point expansion, while the embeddings $\{e^{(d)}_m\}_{m=1}^{k(d)}$ in stage $d$ are shared among all expansion operations of the points in $\mathbb{S}^{(d)}$ (where $e^{(d)}_m \in \mathbb{R}^{64\times 1}$ in our experiments). The initial value for the scaling factor $\alpha^{0}$ at the stage $0$ is set to $1$. The batch size is set to $64$, and we use the AdamW~\cite{loshchilov2017decoupled} optimizer with learning rate
$1\mathrm{e}{-3}$. All
our source code and models will be publicly
available upon paper acceptance.}

\section{Experiments}

\walon{We first introduce the dataset and the evaluation metrics used in our experiments. We then provide the experimental results as well as the comparisons with respect to baseline(s), on tasks of point cloud reconstruction, generation, part segmentation, and co-segmentation.}

\paragraph{Datasets.} 
\walon{The experiments and evaluations are conducted on ShapeNet~\cite{Chang:2015:SAI}, which is a large scale dataset composed of 51,127 pre-aligned 3D shapes from 55 categories. We follow the experimental setting as~\cite{ShapeGF,Yang_2019_ICCV} to have 35,708/5,158 shapes for the train/test split of ShapeNet. Since the shapes in ShapeNet are originally in the format of meshes, we follow the common practice as~\cite{Qi_2017_CVPR} 
to sample random points on the triangles from the meshes in order to build up the point cloud data of each shape for our experiments, where the number of points in each point cloud data is 2048 unless stated otherwise. All the point clouds are zero-centered (i.e. aligning the point cloud center to the origin) and normalized to be within the unit sphere (i.e. coordinates re-scaled by dividing with the distance between the farthest point and the point cloud center).}

\paragraph{Evaluation metrics.} \walon{We follow the prior works~\cite{ShapeGF,Yang_2019_ICCV} to adopt the symmetric Chamfer Distance (CD) for evaluating the performance of point cloud reconstruction. Regarding the point cloud generation, we also follow~\cite{ShapeGF,Yang_2019_ICCV} to use the Minimum Matching Distance (MMD), Coverage (COV), and 1-Nearest Neighbor Accuracy (1-NNA) as metrics to access the quality of generated point clouds. While taking the test split of ShapeNet as the reference set, MMD measures the averaged Chamfer distances from each reference point cloud to its nearest neighbor in the set of generated point clouds, COV measures the fraction of reference point clouds which are matched to at least one sample in the set of generated point clouds, while 1-NNA mainly considers the similarity between the reference set and the generated point clouds in terms of their shape distributions. Note that MMD and 1-NNA are the smaller the better, while COV is instead the larger the better.}

\paragraph{Our RPG variants.} \walon{Two variants are adopted for our RPG training. The first one, denoted as RPG$^{3125}$, is trained on the point cloud data with having 3125 points per point cloud, where there are 5 expansion stages with setting all $k(d)$ equally to 5. Another one, denoted as RPG$^{2048}$, is instead based on the points clouds of 2048 points, where there are also 5 expansion stages but setting $k(1)$ to 8 and $k(2)=k(3)=k(4)=k(5)$ to 4. We mainly adopt RPG$^{2048}$ for evaluating the tasks of reconstruction and generation in order to have fair comparison to the baselines, while using RPG$^{3125}$ to illustrate the part segmentation for better visualization (where the segmentation into $k(1)=5$ parts is easier for clear colorization of points).}

\subsection{Reconstruction}

\begin{figure}[h!]
    \centering
    \includegraphics[width=\columnwidth]{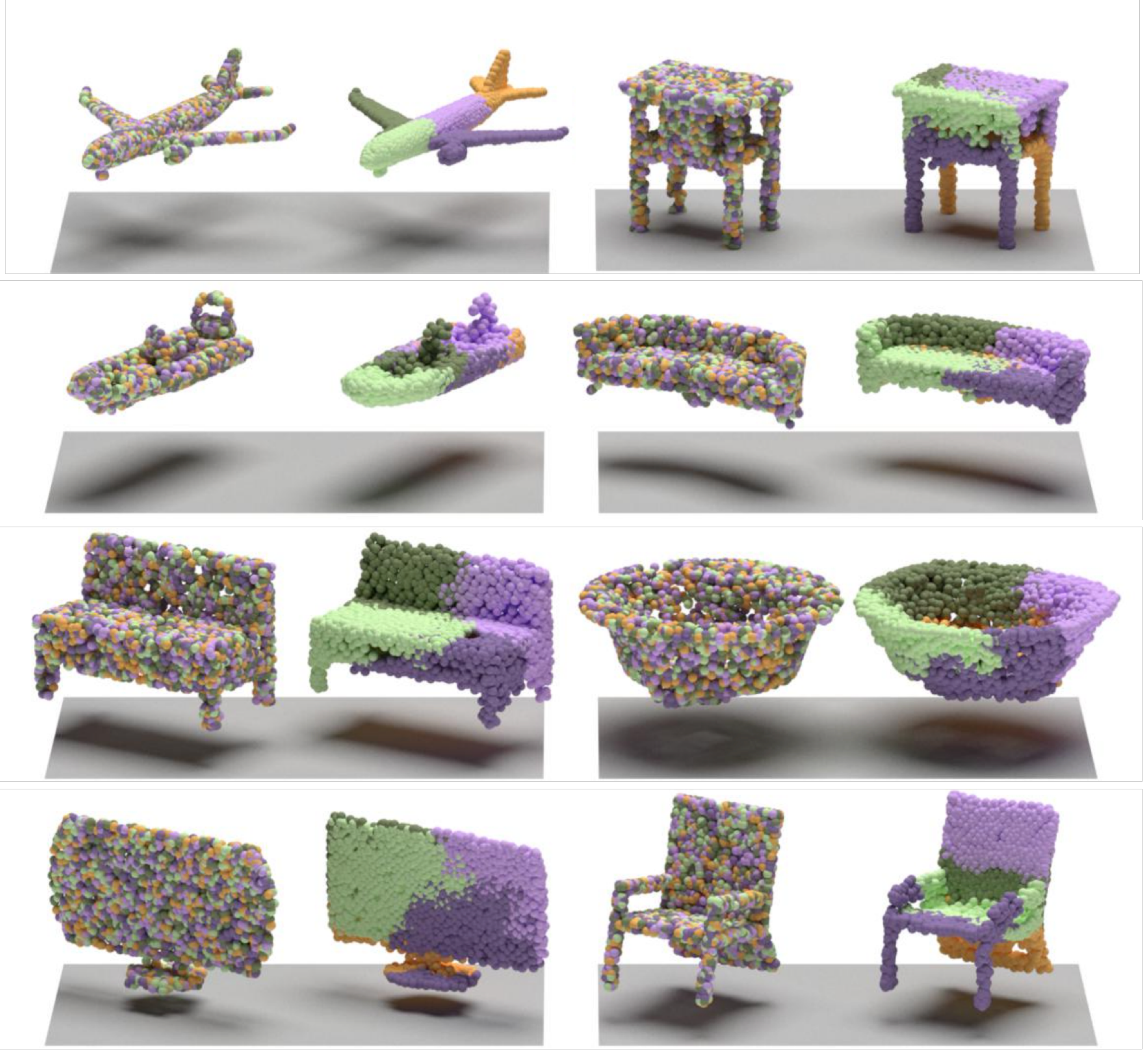}
    \vspace{-1em}
    \caption{\walon{Examples of reconstruction by our RPG. For each pair, the shape on the left is the input while the output is on the right.}}
    \label{fig:reconstruction}
    \vspace{-1em}
\end{figure}

\walon{
We compare the reconstruction ability of auto-encoding the point cloud data among our proposed RPG framework and various baselines, i.e. AtlasNet~\cite{Groueix_2018_CVPR} with patches and with sphere, PointFlow~\cite{Yang_2019_ICCV} (abbreviated as PF), and the current state-of-the-art ShapeGF~\cite{ShapeGF}, based on the whole test set of ShapeNet as well as the three categories of it (i.e. Airplane, Car, and Chair). Please note that, we follow the same experimental setting as~\cite{Yang_2019_ICCV,ShapeGF} where the number of points in each point cloud used for evaluation is 2048. 
The reconstruction errors in terms of Chamfer distance as well as the number of parameters for each method are provided in Table \ref{fig:rec_table}. Note here the performance of our RPG framework is based on RPG$^{2048}$, where the performance obtained by RPG$^{3125}$ is almost the same. It is shown that our proposed RPG framework has the comparable performance of point cloud reconstruction with respect to all the baselines, while having only 1.8M model parameters (with 0.7M from the PointNet encoder $E$), being almost half of the current state-of-the-art ShapeGF~\cite{ShapeGF}. In particular, although the PointFlow~\cite{Yang_2019_ICCV} baseline has smaller number of model parameters than ours, but our reconstruction performance is superior to it on all the datasets. Qualitative examples of reconstruction produced by our RPG are provided in Figure~\ref{fig:reconstruction}.}

\begin{table}[h!]
\caption{\walon{Experimental results of the reconstruction task conducted on the ShapeNet dataset and three categories of it. The reconstruction errors between the input and reconstructed output are given in Chamfer distance multiplied by $10^4$.}}
\label{fig:rec_table}
\resizebox{1\columnwidth}!{
\setlength{\tabcolsep}{0.5mm}{
\begin{tabular}[t]{c|cccccc}
\hline
& \multicolumn{2}{c}{AtlasNet~\cite{Groueix_2018_CVPR}} & \multirow{2}{*}{PF~\cite{Yang_2019_ICCV}} & \multirow{2}{*}{ShapeGF~\cite{ShapeGF}} & \multirow{2}{*}{Ours}\\
\cline{2-3}
Dataset & Sphere & Patches \\
\hline\hline
Airplane & 1.002 & 0.969 & 1.208 & 0.96 & 0.924 \\
\hline
Chair & 6.564 & 6.693 & 10.120 & 5.599 & 7.493 \\
\hline
Car & 5.392 & 5.441 & 6.531 & 5.328 & 5.711  \\
\hline
ShapeNet & 5.301 & 5.121 & 7.551 & 5.154 & 5.607 \\
\hline\hline
Params & 12M & 12M & 1.5M & 3.5M & 1.8M \\
\hline
\end{tabular}
}
}
\end{table}

\subsection{Generation}
\walon{We follow the experimental setting as~\cite{ShapeGF} to conduct the evaluation in terms of generation ability on two categories (i.e. Airplane and Chair) in ShapeNet. Several baselines are adopted for making comparison, including r-GAN~\cite{pmlr-v80-achlioptas18a}, GCN~\cite{valsesia2018learning}, TreeGAN~\cite{Shu_2019_ICCV}, PF~\cite{Yang_2019_ICCV}, and ShapeGF~\cite{ShapeGF}. In order to equip our RPG framework with the ability of generation, our model training is extended from the autoencoder to the variation autoencoder (VAE~\cite{kingma2013auto}), where the distribution of latent representations of 3D shapes are modelled as a normal distribution, thus our RPG can take noise vectors sampled from such normal distribution as $\mathbb{H}^{(0)}$ and output the generated point clouds. The quantitative results of generation quality for various approaches as well as their model sizes are provided in Table~\ref{fig:gen_table}. Please note here we report the performance of our RPG$^{2048}$, which has almost the same performance as RPG$^{3125}$. We can see that, our proposed RPG framework achieves superior performance with respect to r-GAN, GAN, TreeGAN, and PF on almost all the metrics for both categories. While our RPG has comparable performance with ShapeGF, the amount of parameters of our RPG model is smaller than all the baselines thus contributing to less training time and lower hardware requirements. We also provide the qualitative examples of point clouds generated by our proposed RPG in Figure~\ref{fig:gen} to visualize their quality as well as the diversity. Moreover, in Figure~\ref{fig:interpolation}(b) we provide the 5 example sets of interpolation between two 3D shapes of the same category (i.e. shapes generated from the interpolation between two noise vectors which can lead to the point clouds of the same category), where we can see the smooth and semantically-consistent transition between generated point clouds during the interpolation. In Figure~\ref{fig:interpolation}(a) we visualize the point clouds generated on all the expansion stages for interpolated shapes. While in Figure~\ref{fig:interpolation}(c) we provide even the examples of interpolation between different object categories.}
\begin{table}[h!]
\caption{\walon{Experimental results of the generation task conducted on the Airplane and Chair categories of the Shape dataset. Three metrics (i.e. MMD, COV, and 1-NNA) are used to access the quality of generated point clouds, where both MMD and 1-NNA are the lower the better, while COV is the higher the better. The MMD numbers are multiplied with $10^4$ here.}}
\resizebox{1\columnwidth}!{
\setlength{\tabcolsep}{0.75mm}{
\begin{tabular}[t]{c||c|c|ccccc}

\hline
 & Model & Params& MMD($\downarrow$) && COV ($\uparrow$) && l-NNA($\downarrow$) \\
\hline\hline
\multirow{6}{*}{\rotatebox[origin=c]{90}{Airplane}} 
& r-GAN~\cite{pmlr-v80-achlioptas18a} & 6.9M & 1.657  && 38.52  && 95.80 \\ %
& GCN~\cite{valsesia2018learning} & 7.0M & 2.623 && 9.38  && 95.16\\ %
& TreeGAN~\cite{Shu_2019_ICCV} & 40M & 1.466 && 44.69  && 95.06 \\ 
& PF~\cite{Yang_2019_ICCV} & 1.5M & 1.408  && 39.51  && 83.21 \\ %
& ShapeGF~\cite{ShapeGF} & 3.5M & 1.285 && 47.65 && 85.06 \\ %
\cline{2-8}
& Ours & 1.1M & 1.288 && 42.47  && 84.57\\
\hline\hline
\multirow{6}{*}{\rotatebox[origin=c]{90}{Chair}} & r-GAN~\cite{pmlr-v80-achlioptas18a} & 6.9M & 18.187 && 19.49 && 84.82 \\
& GCN~\cite{valsesia2018learning} & 7.0M & 23.098  && 6.95  && 86.52 \\
& TreeGAN~\cite{Shu_2019_ICCV} & 40M & 16.147  && 40.33  && 74.55 \\
& PF~\cite{Yang_2019_ICCV} & 1.5M & 15.027  && 40.94  && 67.60 \\
& ShapeGF~\cite{ShapeGF} & 3.5M  & 14.818 && 46.37 && 66.16 \\
\cline{2-8}
& Ours & 1.1M & 15.686  && 45.62  && 66.39  \\
\hline
\end{tabular}

}}
\label{fig:gen_table}
\vspace{-1em}
\end{table}

\subsection{Part Segmentation and Co-Segmentation}
\begin{figure}[h!]
    \centering
    \includegraphics[width=\columnwidth]{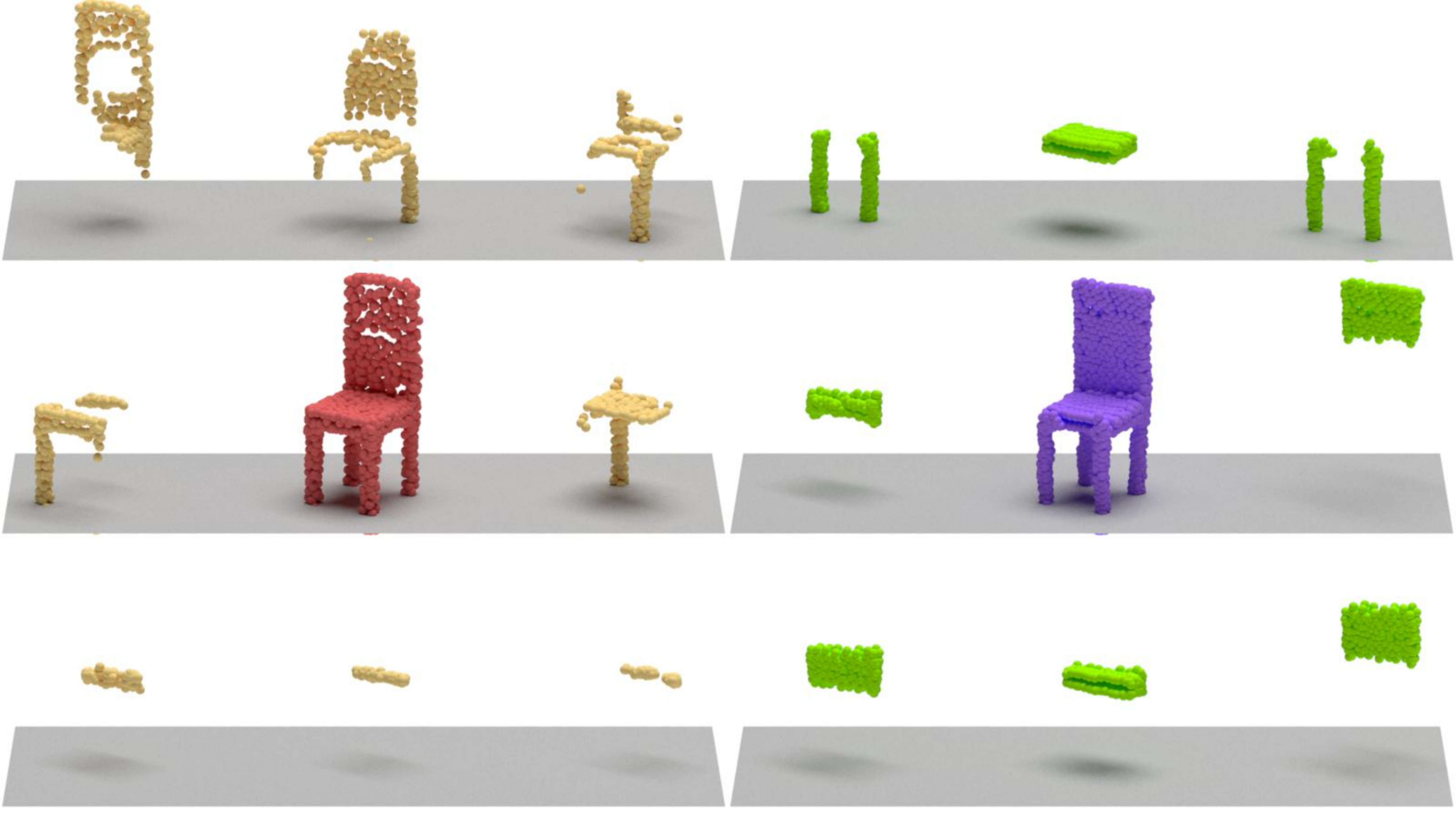}
    \caption{\walon{Comparison between part segmentations obtained from TreeGAN (left) and our proposed RPG (right), where the center image of the left or right part is the generated 3D shape, surrounded by its 8 segmented parts. It shows that our RPG is able to provide more compact and semantically-meaningful parts.}}
    \label{fig:treegan}
    \vspace{-1em}
\end{figure}

\begin{figure*}[h!]
\begin{center}
\includegraphics[width=.96\textwidth]{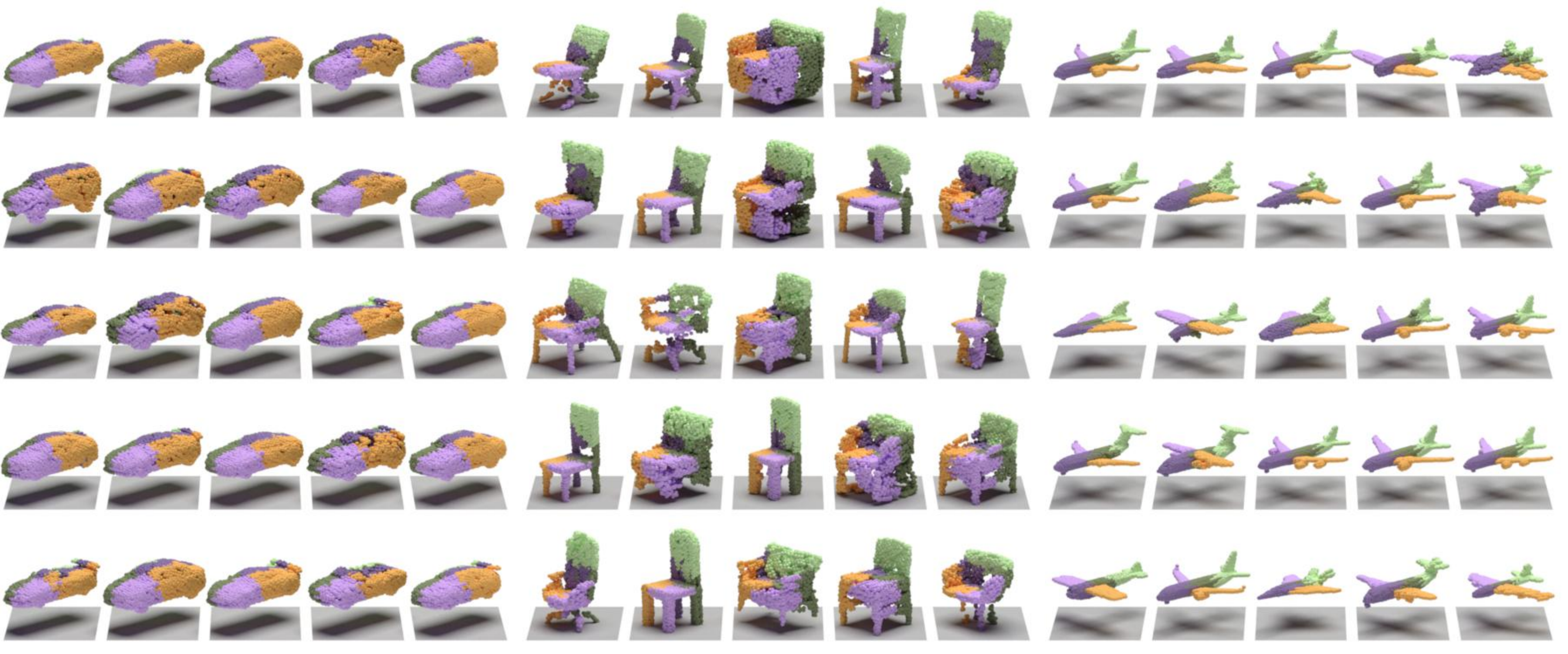}
\end{center}
\vspace{-1.5em}
   \caption{\walon{Qualitative examples of the point clouds generated by our proposed recursive point cloud generator (RPG).}}
\label{fig:gen}
\vspace{-1em}
\end{figure*}

\begin{figure*}[h!]
\begin{center}
\includegraphics[width=.96\textwidth]{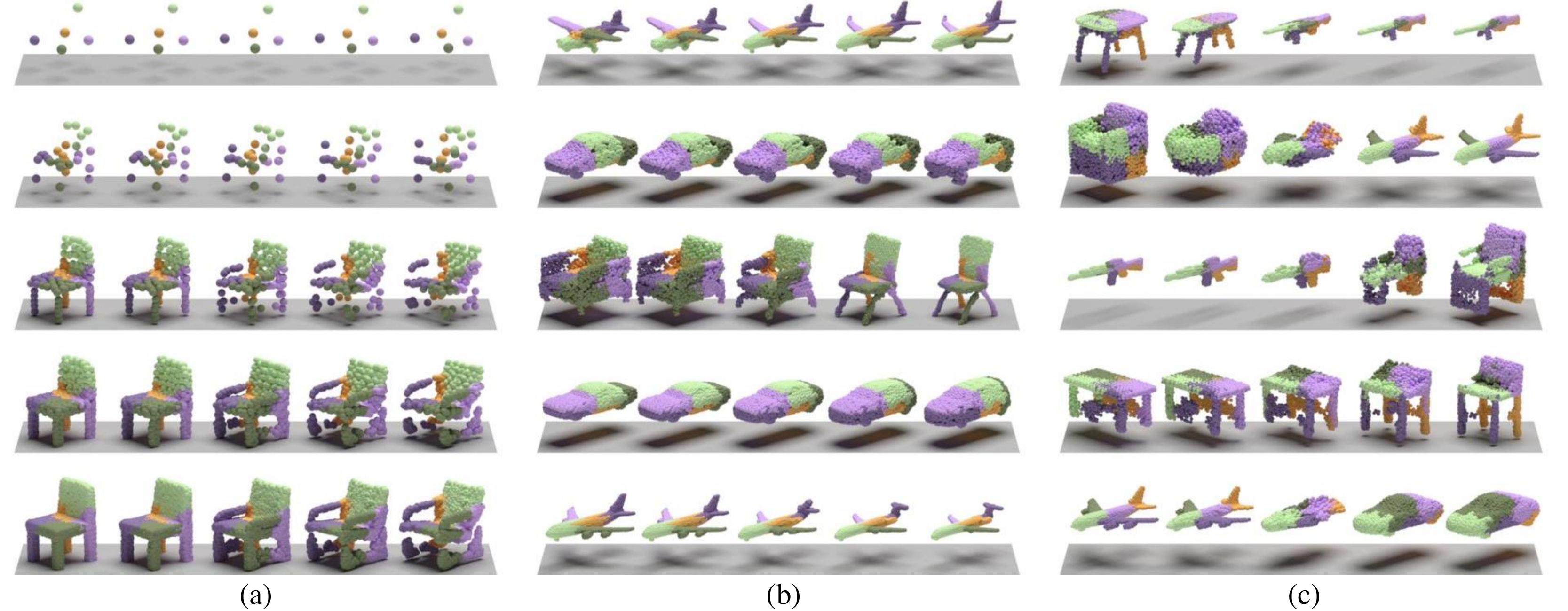}
\end{center}
\vspace{-1.5em}
   \caption{\walon{Examples for our interpolation between different shapes: (a) Rows sequentially show the point clouds generated on all the expansion stages while interpolating between the chairs on the bottom-left and bottom-right corners; (b) Each row shows interpolation between two 3D shapes of the same object category; (c) Each row shows interpolation between two shapes from different categories.}}
\label{fig:interpolation}
\vspace{-1em}
\end{figure*}

\begin{figure*}[h!]
\begin{center}
\includegraphics[width=.96\textwidth]{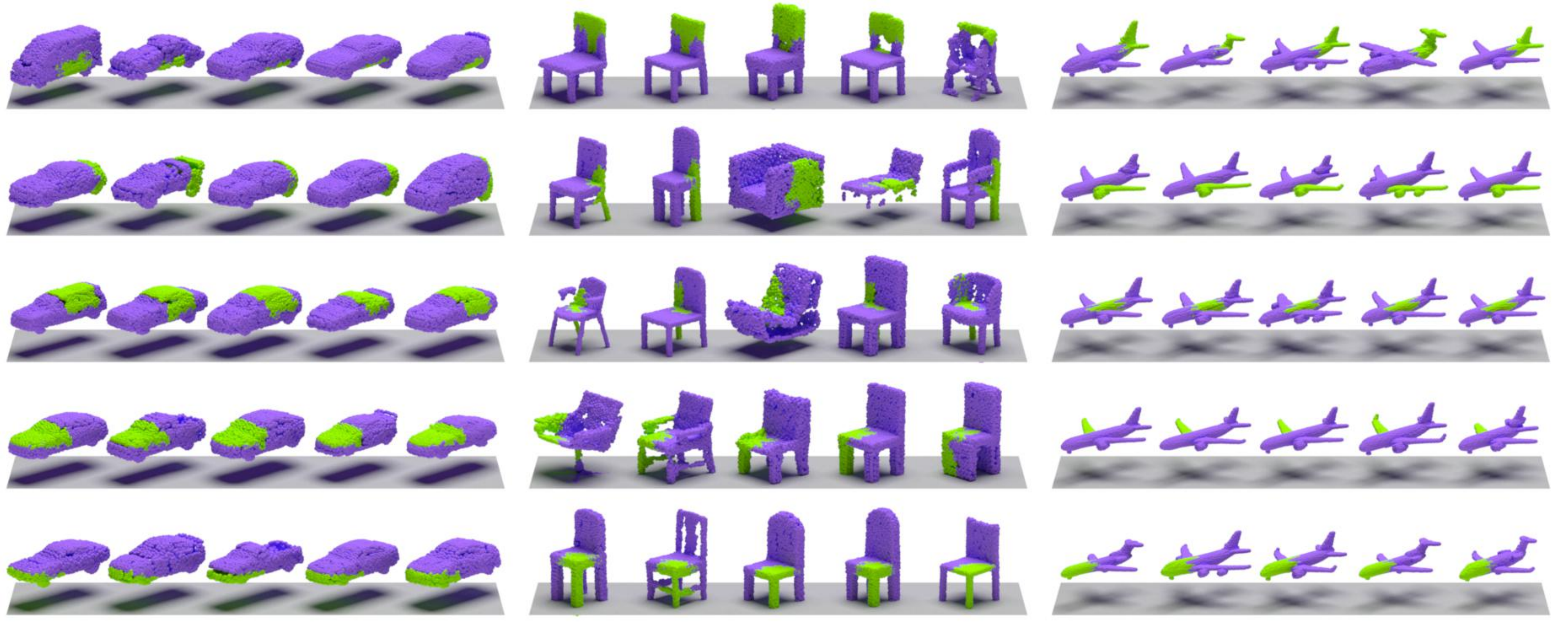}
\end{center}
\vspace{-1.5em}
   \caption{\walon{Visualization of co-segmentation results among object instances from Car, Chair and Airplane categories in ShapeNet. For each object category, the rows sequentially highlight different common parts with green color shared across the instances.}}
\label{fig:co-seg}
\vspace{-1em}
\end{figure*}

\walon{As what can be seen from both Figure~\ref{fig:gen} and~\ref{fig:interpolation}, the points in the generated point clouds are colorized according to their ancestor assignments in $\mathbb{S}^{(1)}$, which particularly highlight the most important feature, i.e. part segmentation, that distinguishing our proposed RPG from the baselines such as PointFlow~\cite{Yang_2019_ICCV} and ShapeGF~\cite{ShapeGF} that can only generate point clouds without having any part segmentation. In particular, the part segmentation is naturally obtained during the inference of our proposed RPG framework, according to the tree-structure built upon the point clouds produced in the sequential stages of expansion (cf. Figure~\ref{fig:tree}) without any supervision nor manual annotation on the object parts. 

Moreover, as the generative procedures of different object instances are learnt by the same recursive process of expansions in our RPG framework, regardless of the tasks of generation or reconstruction, we are therefore able to discover the common patterns of part segmentation shared across object instances thus straightforwardly achieving the co-segmentation, where the common parts (e.g. the left and right wings, tail, and nose of airplanes) are segmented out among objects as shown in Figure~\ref{fig:co-seg}. The co-segmentation helps us to understand the intra-class variance and similarity from a more fine-grained perspective according to the common parts (e.g. different forms of wings among airplanes). 

Furthermore, as TreeGAN proposed in ~\cite{Shu_2019_ICCV} shares several similarities in high-level concepts with our RPG, here we particularly highlight the differences. Basically, TreeGAN adopts the adversarial learning framework to train the generator built upon tree-structured
graph convolution network, where multiple layers of branching operations and the graph convolutions are applied to generate the final point cloud. Due to its tree-structured design, the part segmentation of the generated point cloud is also available. Our work distinguishes from TreeGAN by the following aspects: \textbf{(1)} As the graph convolutions used in different layers of TreeGAN are not shared, its model size is relatively large; while our RPG framework introduces the novel recursive expansion module which is repeatedly applied on all the expansion stages (i.e. weights sharing), the model size of our RPG is much smaller than TreeGAN (cf. Table~\ref{fig:gen_table}) thus leading to more efficient training; \textbf{(2)} Our RPG framework not only provides better performance in generation (cf. Table~\ref{fig:gen_table}) but also shows more compact and semantically-meaningful part segmentation, as shown in Figure~\ref{fig:treegan}. We can see that TreeGAN's part segmentation would mix between the chair back and chair surface (or between chair legs and chair surfaces) or even produce some subtle parts, while ours better decomposes chair legs, surface, and back into different parts thus getting closer to the human interpretation. Moreover, while leveraging the part annotation provided in ShapeNetPart~\cite{yi2016scalable} dataset to evaluate the purity of part segmentation on the point clouds generated by TreeGAN and our RPG, our RPG achieves superior performance than TreeGAN (i.e. 88\% vs. 74\% in purity, please refer to supplementary materials for more details); \textbf{(3)} TreeGAN can only output point cloud on the final layer of generator, while our RPG is able to obtain all the intermediate point clouds thus providing better understanding on the hierarchical segmentation and generative procedure of point cloud data.}

\section{Conclusion}
\walon{In this paper we propose the recursive point cloud generator (RPG) which is capable of generating or reconstructing point clouds in a coarse-to-fine manner. Based on the generative procedure composed of multiple expansion stages, our RPG produces point clouds with semantic parts without requiring any supervision of part annotations. Our proposed RPG achieves comparable performance with respect to state-of-the-art baselines in terms of point cloud generation and reconstruction, with having relatively small model size thanks to the design of sharing weights of the expansion module across all the stages. Moreover, the part segmentation and co-segmentation of point clouds naturally obtained from our RPG provide the potential of understanding point clouds from a more fine-grained perspective.}

{\small
\bibliographystyle{ieee_fullname}
\bibliography{egbib}
}

\end{document}